%% file: main.tex
\documentclass[runningheads]{llncs}
\usepackage[T1]{fontenc}
\usepackage{graphicx}

\usepackage{amsmath}
\usepackage{amssymb}
\usepackage{booktabs}

\usepackage{multirow}
\usepackage{adjustbox}
\usepackage{pifont}
\usepackage{bm}
\usepackage{caption}
\usepackage{color}

\newcommand{\cmark}{\ding{51}}%
\newcommand{\xmark}{\ding{55}}%
\DeclareMathOperator*{\argmax}{argmax}

\def\ie{\emph{i.e}\onedot}

\usepackage[pagebackref,breaklinks,colorlinks]{hyperref}

\usepackage[capitalize]{cleveref}
\crefname{section}{Sec.}{Secs.}
\Crefname{section}{Section}{Sections}
\Crefname{table}{Table}{Tables}
\crefname{table}{Tab.}{Tabs.}
\Crefname{figure}{Figure}{Figures}
\crefname{figure}{Fig.}{Figs.}

\newcommand{\onedot}{.\null}
\newcommand{\etal}{\emph{et al}\onedot}

\def\etal{\emph{et al}\onedot}
\def\eg{\emph{e.g}\onedot}
\def\ie{\emph{i.e}\onedot}

\begin{document}
\title{Surgical Skill Assessment via Video Semantic Aggregation}

\titlerunning{Surgical Skill Assessment via Video Semantic Aggregation}

\author{
Zhenqiang Li\inst{1} \and
Lin Gu\inst{2} \and
Weimin Wang\inst{3} \and
Ryosuke Nakamura\inst{4} \and
Yoichi Sato \inst{1}
}


\authorrunning{Z. Li et al.}

\institute{
The University of Tokyo, Tokyo, Japan \\
    \email{\{lzq, ysato\}@iis.u-tokyo.ac.jp}\\ \and
RIKEN, Tokyo, Japan\\
    \email{lin.gu@riken.jp}\\ \and
Dalian University of Technology, Dalian, China\\
    \email{wangweimin@dlut.edu.cn}\\ \and
AIST, Tsukuba, Japan\\
    \email{r.nakamura@aist.go.jp}
}


\maketitle              

\begin{abstract}
Automated video-based assessment of surgical skills is a promising task in assisting young surgical trainees, especially in poor-resource areas. Existing works often resort to a CNN-LSTM joint framework that models long-term relationships by LSTMs on spatially pooled short-term CNN features. However, this practice would inevitably neglect the difference among semantic concepts such as tools, tissues, and background in the spatial dimension, impeding the subsequent temporal relationship modeling. 
In this paper, we propose a novel skill assessment framework, \textbf{Vi}deo \textbf{S}emantic \textbf{A}ggregation (ViSA), which discovers different semantic parts and aggregates them across spatiotemporal dimensions. The explicit discovery of semantic parts provides an explanatory visualization that helps understand the neural network's decisions. It also enables us to further incorporate auxiliary information such as the kinematic data to improve representation learning and performance. The experiments on two datasets show the competitiveness of ViSA compared to state-of-the-art methods. 
Source code is available at: bit.ly/MICCAI2022ViSA.

\end{abstract}

\input{01_introduction}
\input{03_method}
\input{04_experiments}
\input{05_discussion_conclusion}

\subsubsection{Acknowledgment.}
This work is supported by JST AIP Acceleration Research Grant Number JPMJCR20U1, JSPS KAKENHI Grant Number JP20H04205, JST ACT-X Grant Number JPMJAX190D, JST Moonshot R\&D Grant Number JPMJMS2011, Fundamental Research Funds for the Central Universities under Grant DUT21RC(3)028 and a project commissioned by NEDO.

\bibliographystyle{splncs04}
\bibliography{mainbib}

\end{document}


\author{
Zhenqiang Li\inst{1} \and
Lin Gu\inst{2} \and
Weimin Wang\inst{3} \and
Ryosuke Nakamura\inst{4} \and
Yoichi Sato \inst{1}
}


\authorrunning{Z. Li et al.}

\institute{
The University of Tokyo, Tokyo, Japan \\
    \email{\{lzq, ysato\}@iis.u-tokyo.ac.jp}\\ \and
RIKEN, Tokyo, Japan\\
    \email{lin.gu@riken.jp}\\ \and
Dalian University of Technology, Dalian, China\\
    \email{wangweimin@dlut.edu.cn}\\ \and
AIST, Tsukuba, Japan\\
    \email{r.nakamura@aist.go.jp}
}

\title{Supplementary Material for Surgical Skill Assessment via Video Semantic Aggregation}

\maketitle         
\section{Network Architecture} 

Tab.~\ref{tab:net} provides the architectural details of our ViSA framework. We divide the framework into 8 stages. For each stage, we list its input and output, detailed implementation networks or operations, and output size of each step. 
\input{tex_tables/suptable_arch}

\section{Standard Deviations of Results}
\input{tex_tables/subtable_jigsaws_variance}
We report the standard deviation of the quantitative results on JIGSAWS dataset in \cref{tab:jigsaws_variance}, which is calculated on 6 groups of results with different random seeds. We find that the results on the Leave-one-user-out (LOUO) show higher variances than two other dataset split schemes. We infer that the small number of validation samples in each split causes the large variance. The relatively large value on 4-Fold scheme of Needle Passing task can also be attributed to the few validation samples.
Generally, results on the Leave-one-supertrial-out (LOSO) scheme or the Across task have lower variance. Hence, we think the comparison on the two modes is more stable and reliable. The standard deviations of our experiment results on the Hei-Chole dataset are 0.062 and 0.076 for Corr and MAE metrics respectively.

\section{Visualization of the Results}

\input{tex_images/supvis_numpart}
\subsubsection{Influence of Group Number}
We visualize the assignment maps generated by frameworks with different group number settings for the semantic group module. According to Fig.~\ref{fig:numparts}\redcite{(a)}, video parts are generally divided into groups for robot hands (in cyan) and background (in red) when $K$ is set to 2. When divided into 4 groups, as shown in Fig.~\ref{fig:numparts}\redcite{(c)}, video parts for the hand front ends (in greenyellow), robot arms (in purple), tissue (in red) and background (in cyan) tend to be aggregated to different groups.

\subsubsection{Results by 2D-CNNs}
In Fig.~\ref{fig:r2d}, we illustrate the assignment maps yield by the framework using ResNet-18 as the feature extractor. Akin to the visualization results of the framework based on 3D-CNN features, the 3 groups are also substantially assigned to video parts corresponding to the tool clips (in red), manipulated tissue (in green), and the background (in blue). This implies that our framework, especially the semantic grouping module, can also work well with 2D CNN features. It should also note that the red regions tend to capture the parts of the suture lines in addition to the hand parts. In contrast, by 3D-CNN features, the semantic grouping module always divides the two important parts into two groups (see Fig.~\redcite{2} in the main paper). We infer the reason as the motion information of tool clips embedded in the 3D features helps divide the two parts.
\input{tex_images/supvis_r2d}

\section{Results on TASD-2 Diving Dataset}
\input{tex_tables/res_tasd_bsl}
We also experiment ViSA on the TASD-2 diving dataset~\cite{aqa_asymmetric} for validating its generalizability.
As the quantitative results shown in Tab.~\ref{tab:tasd_baseline}, ViSA outperforms the competitive baseline methods that employ body poses for enhancement.

\input{tex_images/supvis_tasd_numpart}

Fig.~\ref{fig:tasd_numparts} shows the assignment maps of different numbers of groups for one same video. Similar to the assignment maps on the JIGSAWS dataset, ViSA can build the consistent correspondence between the semantic groups and different video parts across space and time. For the 2 groups division, it is obvious that one group corresponds to the background (in red) and another is allocated to the regions (in cyan) the action performers and the splash. According to the rules, both the performance of the action and the amount of splash are primary factors for scoring diving actions. When divided into 3 groups, the regions (in red) around the performers and splash (in green) are separated out from the background (in blue). We think features in these red regions may capture some auxiliary information (\eg, the distance of divers from the platforms or the height of the diver at the apex of the dive) for facilitating the quality assessment of actions. 

\bibliographystyle{splncs04}
\bibliography{mainbib}

%% file: 01_introduction.tex
\section{Introduction}

\label{sec:introduction}
Automated assessment of surgical skills is a promising task in computer-aided surgical training~\cite{surgical_science,surgical_rmis}, especially in the resource-poor countries. Surgical skill assessment involves two major challenges~\cite{surgical_interp}: 1. how to capture the difference between fine-grained atomic actions. 2. how to model the contextual relationships between these actions in the long-term range.
Previous works~\cite{surgical_3dcnn,surgical_bilstm,surgical_unified,surgical_interp,surgical_cnn} counteract  above two challenges by stacking convolutional neural networks (CNNs)  and LSTMs: CNNs for short-term feature extraction and temporal aggregation networks (\eg, LSTMs) for long-term relationship modeling.  
For example, 
Fawaz \etal~\cite{surgical_kinematic} used 1D CNN to encode kinematic data before the aggregation over time by global average pooling. 
Wang \etal ~\cite{surgical_interp} extract 2D or 3D CNN features from video frames and model their temporal relationship  by leveraging an LSTM network. 
Considerable progress has been made in predicting skill levels~\cite{surgical_kinematic,surgical_3dcnn,surgical_bilstm,sd_manipulation,surgical_shmm,surgical_cnn,surgical_accel} or scores~\cite{surgical_unified,surgical_interp,surgical_rmis} using kinematic data or video frames. 

Regarding the video-based methods, most methods~\cite{surgical_3dcnn,surgical_bilstm,surgical_unified,surgical_cnn,surgical_interp} perform the global pooling over the spatial dimension on CNN features before feeding them to the subsequent network. We argue that this global pooling  operation ignores the semantics variance of different features and compresses all the information together in the spatial scale without distinction. As a result, the consequent networks could hardly model the temporal relationship of the local features in different spatial parts separately, \eg, the movements of different tools and the status changes of tissue.
This bottleneck is particularly severe for surgical skill assessment because the tracking of tools from a large part of the background is essential to judge the manipulation quality. Similarly, the interactions between tools and tissue across time are also important for the assessment. To make it worse, most of existing methods are end-to-end neural networks, revealing little about what motion or appearance information is captured.

Since surgical videos comprise limited objects with explicit semantic meanings such as tools, tissue, and background, we propose a novel framework, \textbf{Vi}deo \textbf{S}emantic \textbf{A}ggregation (ViSA), for surgical skill assessment by aggregating local features across spatiotemporal dimensions according to these semantic clues. This aggregation allows our method to separate the video parts related to different semantics from the background and further dedicate to tracking the tool or modeling its interaction with tissue. 

\begin{figure*}[t]
\centering
\includegraphics[width=1.0\textwidth]{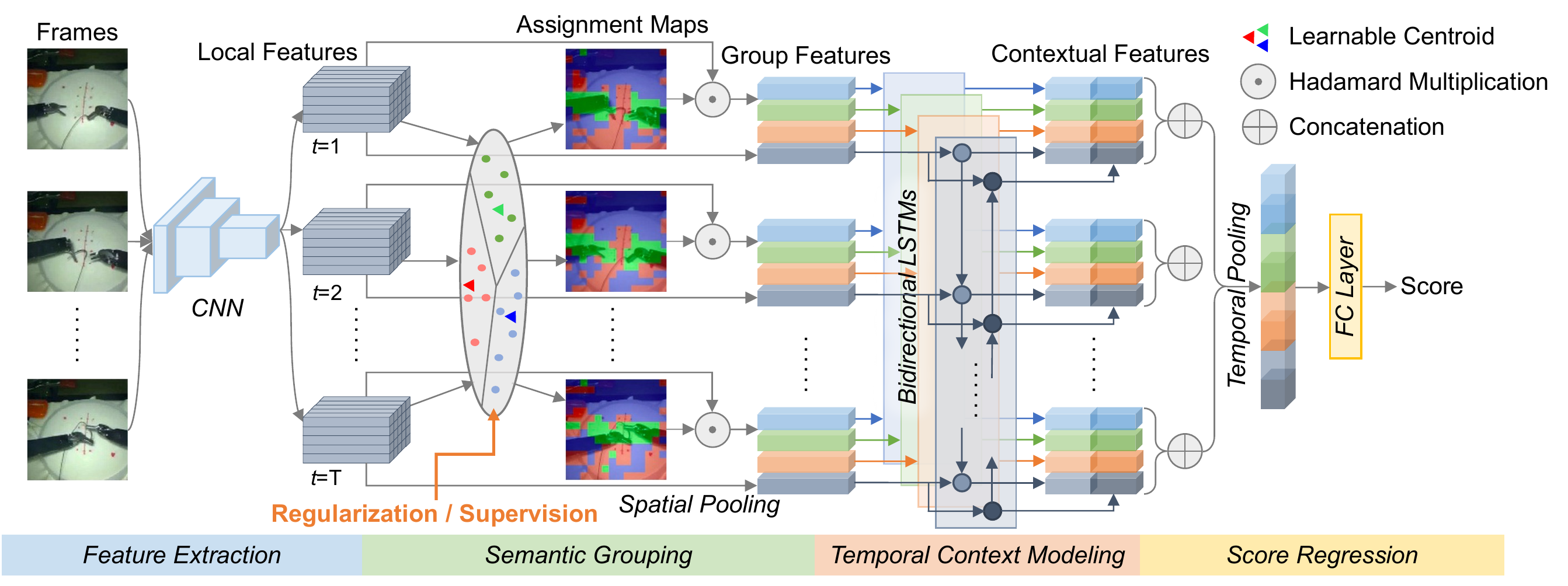}
\caption{The proposed framework, \textbf{Vi}deo \textbf{S}emantic \textbf{A}ggregation (ViSA), for surgical skill assessment. It aggregates local CNN features over space and time using semantic clues embedding in the video. We aim to group video parts according to semantics, \eg, tools, tissue and background, and model the temporal relationship for different parts separately via this framework. We also investigate the regularization or additional supervision to the group results for enhancement.}
\label{fig_framework}
\end{figure*}

As shown in \cref{fig_framework}, ViSA first aggregates similar local semantic features through clustering and generates the abstract features for each semantic group in the semantic grouping stage. Then we aggregate the features for the same semantic across time and model their temporal contextual relationship via multiple bidirectional LSTMs. The spatially aggregated features can visualize the correspondence between different video parts and semantics in the form of assignment maps, facilitating the transparency of the network. In addition, the explicit semantic group of features allows us to incorporate auxiliary information that further improves the assignment and influences the intermediate features, \eg, using kinematic data to bound features for certain semantics to tools. 

Our contribution is threefold:
(1) We propose a novel framework, ViSA, to assess skills in surgical videos via explicitly splitting different semantics in video and efficiently aggregating them across spatiotemporal dimensions.
(2) Via aggregating the video representations and regularization, our method can discover different semantic parts such as tools, tissue, and background in videos. This provides explanatory visualization as well as allows integrating auxiliary information like tool kinematics for performance enhancement.
(3) The framework achieves competitive performance on two datasets: JIGSAWS~\cite{jigsaws} and HeiChole~\cite{heichole}. 

%% file: 03_method.tex
\section{Methodology}

As shown in Fig.~\ref{fig_framework}, for each video, our framework takes frames from a fixed number of timesteps as input and predicts the final skill assessment score through the following 4 stacked modules: (1) the Feature Extraction Module (FEM) that produces feature maps for each timestep (\cref{sec:feature}); (2) the Semantic Grouping Module (SGM) that aggregates local features into a specified number of groups based on the embedded semantics (\cref{sec:region}); (3) the Temporal Context Modeling Module (TCMM) that models the contextual relationship for the feature series of each group (\cref{sec:context}); (4) the Score Prediction Module (SPM) that regresses the final score based on the spatiotemporally aggregated features (\cref{sec:score}).

\subsection{Feature Extraction Module}

\label{sec:feature}
Our framework first feeds the input frames to CNNs to collect the intermediate layer responses as the feature maps for the subsequent processing. We stack the feature maps for one video along the temporal dimension and registered them as $\bm{X}\in\mathbb{R}^{T\times H\times W\times C}$, where $T$ is the number of timesteps, $H, W$ represents the height and width of each feature map, and $C$ is the number of channels. $\bm{X}$ can also be seen as a group of local features, where each feature is indexed by the temporal and spatial position $t, i, j$ as $\bm{X}_{tij}\in\mathbb{R}^C$.

\subsection{Semantic Grouping Module}

\label{sec:region}
This module aggregates the extracted features $\bm{X}$ into a fixed number of groups, with each representing a specific kind of semantic meaning.
Taking inspirations from~\cite{actionvlad,featgroup_regiongroup}, we achieve this by clustering all local CNN features across the entire spatiotemporal range of the video according to $K$ learnable vectors $\bm{D}=[\bm{d}_1,\cdots ,\bm{d}_K]^\top\in\mathbb{R}^{K\times C}$ which record the feature centroid of each group (drawn as colored triangles in \cref{fig_framework}). 

\subsubsection{Local Feature Assignment} 
We softly assign local features to different groups by computing the assignment possibilities $\bm{P}\in\mathbb{R}^{T\times H\times W\times K}$ subject to $\bm{P}^k_{tij} = \frac{\text{exp}(-\parallel(\bm{X}_{tij}-\bm{d}_k)/\sigma_k\parallel^2_2)}{\sum_{k'}\text{exp}(-\parallel(\bm{X}_{tij}-\bm{d}_{k'})/\sigma_{k'}\parallel^2_2)}$.
Component $\bm{P}^k_{tij}$ represents the possibility of assigning the local CNN feature $\bm{X}_{tij}$ to the $k^{\text{th}}$ semantic group. $\sigma_k\in(0,1)$ is a learnable factor to adjust feature magnitudes and smooth the assignment for each semantic group. Taking the index of the group with the maximum possibility at each position, we obtain the assignment maps $\bm{A} \in\mathbb{R}^{T\times H\times W}$ with component
$
\bm{A}_{tij}=\argmax_{k=1,\dots,K}\bm{P}^k_{tij}, \forall t,i,j.
$
As shown in \cref{fig_framework}, the assignment maps visualize the correspondence between video parts and semantic groups. 

\subsubsection{Group Feature Aggregation}
For each timestep, we aggregate the corresponding local features according to the soft-assignment results and generate an abstract representation for each semantic group. We first calculate the normalized residual between the centroid and the averaged local features weighted by assignment possibilities as $\bm{z}^k_t=\frac{\bm{z}^{\prime k}_t}{\parallel \bm{z}^{\prime k}_t\parallel_2}$, where $\bm{z}^{\prime k}_t = \frac{1}{\sigma_k}\big(\sum_{ij}\frac{{\bm{P}^k_{tij}\bm{X}_{tij}}}{\sum_{ij}{\bm{P}^k_{tij}}}-\bm{d_k}\big)$.
Then we get one \textit{group feature} $\bm{g}^k_t$ for capturing the abstract information of the $k^\text{th}$ semantics at each timestep by transforming $\bm{z}^k_t$ with a sub-network $f_g$ consisting of 2 linear transformations as $\bm{g}_t^k = f_g(\bm{z}^k_t)$.

\subsubsection{Group Existence Regularization}
To avoid most local features being allocated to one single group, we leverage a regularization term~\cite{featgroup_regiongroup} to retain the even distribution among groups at every timestep.
Specifically, it regularizes the existence of each group by constraining the max assignment probability of one group, \ie, $\hat{p}_t^k:=\max_{i,j}\bm{P}_{t,i,j}^k$, as close to 1.0 as possible. 
Instead of tightly constraining $\hat{p}_t^k$ to an exact number, the regularization term restricts the cumulative distribution function of $\hat{p}^k$ to follow a beta distribution. We implement the regularization term as $\mathcal{L}_{exist} = \sum_{i=1}^T \Big(\log\big(\bm{p}^{\star k}_i+\epsilon\big)-\log\big(F^{-1}(\frac{2i-1}{2T};\alpha, \beta)+\epsilon\big)\Big)$.
$\bm{p}^{\star k} = \text{sort}([\hat{p}_1^k\cdots \hat{p}_T^k])$
arranges the maximum assignment probabilities of the $k^{\text{th}}$ group at all timesteps in ascending order. $\alpha$=1 and $\beta$=0.001 control the shape of the beta distribution. $\epsilon$ is a small value for numerical stability. 
$F^{-1}(\frac{2i-1}{2T};\alpha, \beta)$ means an inverse cumulative beta distribution function, which returns the $i^{\text{th}}$ element of the sequence of $T$ probability values obeying this distribution.

\subsection{Temporal Context Modeling Module}

\label{sec:context}
For each obtained semantic group feature series, this module aims to keep track of its long-term dependencies and model their contextual relationships independently in a recurrent manner. As shown in \cref{fig_framework}, we achieve this by employing $K$ bidirectional LSTMs (BiLSTMs): $\overrightarrow{\bm{h}}_t^k = \overrightarrow{L}^k(\overrightarrow{\bm{h}}_{t-1}^k,\bm{g}_t^k)$, $\overleftarrow{\bm{h}}_t^k = \overleftarrow{L}^k(\overleftarrow{\bm{h}}_{t+1}^k,\bm{g}_t^k)$,
where $\overrightarrow{L}^k$ and $\overleftarrow{L}^k$ denote the $k^\text{th}$ forward and backward LSTMs respectively.
The output vectors of every timestep $\overrightarrow{\bm{h}}_t^k$ and $\overleftarrow{\bm{h}}_t^k$ from the two directions are further concatenated to form the \textit{contextual feature} $\bm{c}_t^k = [\overrightarrow{\bm{h}}_t^{k\top}, \overleftarrow{\bm{h}}_t^{k\top}]^\top$.
To prevent the potential information loss caused by the separated  modeling (\eg, the interaction between groups), we also employ another BiLSTM to model the global features $[\bm{x}_1\cdots \bm{x}_T]$ which are obtained by taking spatial average pooling on $\bm{X}$. $\bm{c}_t^0$ denotes the generated contextual features by this additional BiLSTM.

\subsection{Score Prediction Module}

\label{sec:score}
This module concatenates the contextual features $\bm{c}_t^k$ of different semantic groups at the same timestep into a vector, followed by an average pooling on these vectors across the temporal dimension. The pooled vector is finally regressed to the skill score $s$ by $f_s$, a fully connected layer. We formulate this module mathematically as follows: $\hat{s} = f_s(\frac{1}{T}\sum_{t=1}^{T}[\bm{c}_t^{0\top}\cdots \bm{c}_t^{K\top}]^\top)$.
The training loss function is defined as:
$
    \mathcal{L} = (s - \hat{s})^2 + \lambda\mathcal{L}_{exist},
$
with $s$ denoting the ground-truth score.
The scalar $\lambda=10$ controls the contribution of the group existence regularization.

\subsection{Incorporating Auxiliary Supervision Signals}

\label{sec:supervision}
Our unique representation aggregation strategy also allows enhancing the feature grouping results by incorporating additional supervision signals to assign specific known semantic information to certain groups. Specifically, we propose a Heatmap Prediction Module (HPM), which takes as input the CNN feature maps $\bm{X}$ and the assignment possibility maps $\bm{P}^m$ of the $m^{\text{th}}$ group. The module predicts the positions of the specified semantics by generating heatmaps $\hat{\bm{H}}=f_{HPM}(\bm{P}^m \odot \bm{X}) \in\mathbb{R}^{T\times H\times W}$,
where $\odot$ denotes the Hadamard product, and $f_{HPM}$ is composed of one basic $1\times1$ conv block followed by a Sigmoid function. Assuming the 2D positions of the specified semantics is known, we generate the position heatmaps $\bm{H}\in\mathbb{R}^{T\times H\times W}$ from the 2D positions as the supervision signal. The framework integrating HPM is trained by the loss function with an extra position regularization $\mathcal{L'} = (s - \hat{s})^2 + \lambda_1\mathcal{L}_{exist}+\lambda_2\mathcal{L}_{pos}$,
where $\mathcal{L}_{pos}$ computes the binary cross entropy between $H$ and $\hat{H}$ and $\lambda_1$=10, $\lambda_2$=20.

%% file: 04_experiments.tex
\section{Experiments}

\input{tex_tables/res_jigsaws_bsl}

\noindent\textbf{Dataset} We evaluate ViSA framework on 2 datasets for surgical skill assessment: JIGSAWS~\cite{jigsaws} and HeiChole~\cite{heichole}. JIGSAWS is a widely used dataset consisting of 3 elementary surgical tasks: Knot Tying (KT), Needle Passing (NP), and Suturing (SU). 
Each task contains more than 30 trials performed on the da Vinci surgical system, which is rated from 6 aspects. Following~\cite{surgical_interp,surgical_unified,surgical_rmis}, the sum score is used as the ground truth. 
We validate our method on every task by 3 kinds of cross-validation schemes: Leave-one-supertrial-out (LOSO), Leave-one-user-out (LOUO) and 4-Fold~\cite{aqa_jointgraph,aqa_uncertainty,aqa_asymmetric}. 
HeiChole is a challenging dataset containing 24 endoscopic videos in real surgical environments for laparoscopic cholecystectomy. For each video, 2 clips of the phases calot triangle dissection and gallbladder dissection are provided with skill scores from 5 domains. We use the sum score as the ground truth. We train and validate our framework on the 48 video clips by 4-fold cross-validation with a 75/25 partition ratio.

\noindent\textbf{Evaluation Metric} For JIGSAWS, since previous works~\cite{surgical_interp,surgical_unified} only included the results on Spearman's Rank Correlation (\textbf{Corr}), we report the validation correlations averaged on all folds for one task and compute the average correlation across tasks by Fisher’s z-value~\cite{aqa_multiactions} for baseline comparison. We incorporate the results on Mean Absolute Error (\textbf{MAE}) in ablation studies. For HeiChole, we report both the correlation and MAE averaged on all folds. The reported results are averaged on multiple runs with different random seeds.

\noindent\textbf{Implementation Details} 
We employ R(2+1)D-18~\cite{r2p1d} pre-trained on Kinetic-400 dataset~\cite{kinetics} as the feature extractor and take the response of the 4$^\text{th}$ convolutional block as the 3D spatiotemporal feature. Since each 3D feature is extracted from 4 frames, we divide each video into $T$ segments and sample a 4-frame snippet from each. $T$ is set to 32 on JIGSAWS videos and 64 on longer HeiChole videos.
Each sampled frame is resized to $160\times120$ and crop the $112\times112$ region in the center as input. Hence, the spatial size of the extracted CNN feature maps is $14\times14$. We also investigate 2D-CNNs feature extractor by using ResNet-18 pre-trained on ImageNet.
Since the surgical scenes in our experiments explicitly include 3 parts: represented by tools, tissues, and background, we initialize the number of semantic groups $K=3$. It should be set according to the scene complexity and fine-tuned based on the experiment results.
The framework is implemented by PyTorch. We use SDG optimizer with mean squared error loss. Models are trained in 40 epochs with a batch size of 4. Learning rates are initialized as 3e-5 and decayed by 0.1 times for every 20 epochs.

\input{tex_tables/res_heichole_bsl_jigsaws_abl}

\subsection{Baseline Comparison}

\cref{tab:jigsaws_baseline} shows the quantitative baseline comparison results of ViSA on JIGSAWS. Baseline results are taken from previous papers. We find that ViSA outperforms many competitive methods on most tasks and cross-validation schemes. 
ViSA has a CNN-RNN framework akin to C3D-MTL-VF~\cite{surgical_interp} but surpasses it with obvious margins on most LOSO and LOUO metrics. 
On 4-Fold, ViSA  achieves nearly equivalent performance as MultiPath-VTPE~\cite{surgical_unified} which needs extra input sources such as frame-wise surgical gestures information and tool movement paths. 
For HeiChole, since no baseline is available, we newly constructed four basic frameworks by combining 2D or 3D CNN feature extractors with temporal aggregation modules of fully connected layers (FC) or LSTMs, which share similar CNN architectures and pre-trained parameters as ViSA. \cref{tab:heichole_baseline} shows the out-performance of ViSA against the four baselines on both metrics.

\input{tex_images/vis_figure}
\input{tex_images/vis_htmpregu_gradcam}

\subsection{Assignment Visualization}

\cref{fig_assignment} visualizes the assignment maps generated by ViSA on two datasets. 
The three semantic groups generally correspond to the tools, the manipulated tissues, and the background regions as expected. 
Notably, ViSA gets this assignment result without taking any supervision of semantics, which further indicates its effectiveness in modeling surgical actions over spatiotemporal dimensions.

\subsection{Ablation Study}

We also conduct the ablation analysis on three key components and one hyper-parameter of ViSA: Feature Extraction Module (\textbf{FEM}), Semantic Grouping Module (\textbf{SGM}), Temporal Context Modeling Module (\textbf{TCMM}), and the number of semantic groups (\textbf{K}). Referring to \cite{surgical_interp}, we train and validate the ablative frameworks across videos of 3 JIGSAWS tasks by forming them together, in order for stable results on more samples. Results in \cref{tab:jigsaws_ablation} indicate that: (1) leveraging SGM boosts the performance on either R(2+1)D-18 or R2D-18; (2) BiLSTMs perform better than LSTMs and Transformer~\cite{transformer} in modeling temporal context; (3) increasing $K$ from 2 to 3 causes more improvements than raising it from 3 to 4, which indicates separating semantics into 3 groups fits the JIGSAWS dataset. Transformer consists of two layers of LayerNorm+Multi-Head-Attention+MLP. Considering the data-hungry nature of Transformer~\cite{datahungrytransformer}, we attribute its unremarkable performance to the small amount of training data.

\cref{fig:gradcam} presents the visual explanation results generated by the post-hoc interpretation method Grad-CAM~\cite{video_attribution,gradcam} which localizes the input regions used by networks for decision making. Compared to the network without using the semantic grouping module (SGM), the full framework employs more task-related parts for predictions (\eg, regions about robotic tools) and discards many unrelated regions. We attribute the improvement to explicitly discovering different video parts and modeling their spatiotemporal relationship in our framework.

\subsection{Improved Performance with Supervision}


ViSA also supports the explicit supervision on the grouping process that allocates specific semantics to certain expected groups. On JIGSAWS, the kinematic data recording the positions of two robotic clips in 3D space is available. 
We first approximate the two clips' positions on the 2D image plane by projecting their 3D kinematic positions with estimated transformation matrix. Then we generate the heatmaps $\bm{H}\in\mathbb{R}^{T\times H\times W}$ from the 2D positions and train the framework as described in \cref{sec:supervision}. 
\cref{fig:heatmap_regu} illustrates one example of the generated position heatmaps. 
On the Suturing task, we achieve the improvement in average validation correlations as $0.84\rightarrow 0.87$, $0.72\rightarrow 0.80$ and $0.80\rightarrow 0.86$ on LOSO, LOUO, and 4-Fold schemes respectively. In \cref{fig:heatmap_regu}, we show one imperfect assignment maps generated by the model without supervision and its corresponding maps after using the position supervision. 
Although the position supervision is not fully precise, it still benefits the network to discover the tools' features. Hence, in \cref{fig:heatmap_regu}, the green regions become less noisy and are guided to the regions of tool clips and tool-tissue interactions after using the supervision.


%% file: tex_tables/res_jigsaws_bsl.tex
\begin{table*}[t]
\centering
\caption{Baseline comparison on JIGSAWS. We report the Spearman's Rank Correlations by three cross-validation schemes on every task. \textbf{K}: Kinematic data, \textbf{V}: video frames, *: extra annotations such as surgical gestures are utilized.}
\label{tab:jigsaws_baseline}
\begin{adjustbox}{width=\textwidth}
\setlength{\tabcolsep}{4pt}
\begin{tabular}{@{}ll|ccc|ccc|ccc|ccc@{}}
\toprule
\multirow{3}{*}{Input} & \multirow{3}{*}{Method} & \multicolumn{12}{c}{Task \& Scheme} \\ 
\cmidrule{3-14} 
 &  & \multicolumn{3}{c}{KT} & \multicolumn{3}{c}{NP} & \multicolumn{3}{c}{SU} & \multicolumn{3}{c}{Avg.} 
\\ 
 \cmidrule(lr){3-5} \cmidrule(lr){6-8} \cmidrule(lr){9-11} \cmidrule(lr){12-14} 
 &  & LOSO & LOUO & 4-Fold & LOSO & LOUO & 4-Fold & LOSO & LOUO & 4-Fold & LOSO & LOUO & 4-Fold \\ 
\midrule
\multirow{2}{*}{\textbf{K}}
 & SMT-DCT-DFT~\cite{surgical_rmis} & 0.70 & 0.73 & - & 0.38 & 0.23 & - & 0.64 & 0.10 & - & 0.59 & 0.40 & - \\
 & DCT-DFT-ApEn~\cite{surgical_rmis} & 0.63 & 0.60 & - & 0.46 & 0.25 & - & 0.75 & 0.37 & - & 0.63 & 0.41 & -\\ 
\midrule

\multirow{9}{*}{\textbf{V}} 
 & ResNet-LSTM~\cite{surgical_interp} & 0.52 & 0.36 & - & 0.84 & 0.33 & - & 0.73 & 0.67 & - & 0.72 & 0.59 & - \\
 & C3D-LSTM~\cite{aqa_olympic} & 0.81 & 0.60 & - & 0.84 & 0.78 & - & 0.69 & 0.59 & - & 0.79 & 0.67 & - \\
 & C3D-SVR~\cite{aqa_olympic} & 0.71 & 0.33 & - & 0.75 & -0.17 & - & 0.42 & 0.37 & - & 0.65 & 0.18 & - \\ 
 & USDL~\cite{aqa_uncertainty} & - & - & 0.61 & - & - & 0.63 & - & - & 0.64 & - & - & 0.63 \\
 
 & MUSDL~\cite{aqa_uncertainty} & - & - & 0.71 & - & - & 0.69 & - & - & 0.71 & - & - & 0.70 \\
 & *S3D~\cite{aqa_s3d} & 0.64 & 0.14 & - & 0.57 & 0.35 & - & 0.68 & 0.03 & - & - & - & - \\
 & *ResNet-MTL-VF~\cite{surgical_interp} & 0.63 & 0.72 & - & 0.73 & 0.48 & - & 0.79 & 0.68 & - & 0.73 & 0.64 & - \\
 & *C3D-MTL-VF~\cite{surgical_interp} & 0.89 & \textbf{0.83} & - & 0.75 & 0.86 & - & 0.77 & 0.69 & - & 0.75 & 0.68 & - \\ 
\midrule

\multirow{5}{*}{\textbf{V}+\textbf{K}} 
 & JR-GCN~\cite{aqa_jointgraph} & - & 0.19 & 0.75 & - & 0.67 & 0.51 & - & 0.35 & 0.36 & - & 0.40 & 0.57 \\
 & AIM~\cite{aqa_asymmetric} & - & 0.61 & 0.82 & - & 0.34 & 0.65 & - & 0.45 & 0.63 & - & 0.47 & 0.71 \\
 & MultiPath-VTP~\cite{surgical_unified} & - & 0.58 & 0.78 & - & 0.62 & 0.76 & - & 0.45 & 0.79 & - & 0.56 & 0.78 \\
 & *MultiPath-VTPE~\cite{surgical_unified} & - & 0.59 & 0.82 & - & 0.65 & 0.76 & - & 0.45 & \textbf{0.83} & - & 0.57 & 0.80 \\
\midrule

\multirow{1}{*}{\textbf{V}} 
 & ViSA & \textbf{0.92} & 0.76 & \textbf{0.84} & \textbf{0.93} & \textbf{0.90} & \textbf{0.86} & \textbf{0.84} & \textbf{0.72} & 0.79 & \textbf{0.90} & \textbf{0.81} & \textbf{0.83} \\


\bottomrule
\end{tabular}
\end{adjustbox}
\end{table*}

%% file: tex_tables/res_heichole_bsl_jigsaws_abl.tex
\begin{table}[t!]
\centering
\begin{minipage}{0.34\textwidth}
\centering
\caption{Results on HeiChole. Baseline frameworks are newly constructed and compared with ViSA by MAE and Corr metrics.}
\label{tab:heichole_baseline}
\setlength{\tabcolsep}{4pt}
\begin{adjustbox}{width=\linewidth}

\begin{tabular}{l|c|c}
\toprule
Method & MAE & Corr \\ 
\midrule
R2D-18 + FC & 1.56 & 0.32 \\
R2D-18 + LSTM & 1.42 & 0.15 \\
R(2+1)D-18 + FC & 1.54 & 0.29 \\
R(2+1)D-18 + LSTM & 1.33 & 0.31 \\
\midrule
ViSA (R2D-18) & \textbf{1.27} & \textbf{0.46} \\
ViSA (R(2+1)D-18) & \textbf{1.27} & \textbf{0.46} \\
        
\bottomrule
\end{tabular}
\end{adjustbox}
\end{minipage}
\hspace{0.5em}
\begin{minipage}{0.60\textwidth}
\centering
\caption{Ablation studies on JIGSAWS. We train frameworks across videos of three tasks and report Corr and MAEs on LOSO and LOUO settings. K denotes the number of Groups.}
\label{tab:jigsaws_ablation}
\setlength{\tabcolsep}{4pt}
\begin{adjustbox}{width=0.8\linewidth}
\newcommand{\csp}{\hskip 0.1in}
\begin{tabular}{c@{\csp}c@{\csp}c@{\csp}c@{\csp}c|c|c|c}
\toprule
\multirow{2}{*}{FEM} &
\multirow{2}{*}{SGM} &
\multirow{2}{*}{TCMM} &
\multirow{2}{*}{K} &
\multicolumn{2}{c}{LOSO} &
\multicolumn{2}{c}{LOUO} \\ 
\cmidrule(lr){5-6} \cmidrule(lr){7-8} 
& & & & MAE & Corr & MAE & Corr \\
\midrule
R2D-18 & \xmark  & BiLSTMs & - & 3.40 & 0.65 & 3.53 & 0.67 \\
R2D-18 & \cmark  & BiLSTMs & 3 & 3.27& 0.80 & 3.42 & 0.74 \\
\midrule
R(2+1)D-18   & \cmark  & LSTMs   & 3 & 2.41 & 0.83 & 3.23 & 0.78 \\
R(2+1)D-18   & \cmark  & Transformer & 3 & 3.02 & 0.76 & 3.27 & 0.68 \\
\midrule
R(2+1)D-18   & \xmark  & BiLSTMs & - & 2.90 & 0.73 & 3.30 & 0.72 \\
R(2+1)D-18   & \cmark  & BiLSTMs & 2 & 2.34 & 0.84 & 3.07 & 0.72 \\
R(2+1)D-18   & \cmark  & BiLSTMs & 4 & 2.32 & 0.85 & \textbf{2.68} & \textbf{0.79} \\
R(2+1)D-18   & \cmark  & BiLSTMs & 3 & \textbf{2.24} & \textbf{0.86} & 2.86 & 0.76 \\
\bottomrule
\end{tabular}
\end{adjustbox}
\end{minipage}
\end{table}

%% file: tex_images/vis_figure.tex
\begin{figure*}[t!]
	\centering
	\footnotesize
	\renewcommand{\tabcolsep}{1pt}
	\newcommand{\sz}{0.1}
	\begin{tabular}{ccccc|ccccc}
	\multicolumn{4}{l}{\scriptsize{JIGSAWS}} &  &  & \multicolumn{4}{l}{\scriptsize{HeiChole}} \\
	
	\includegraphics[height=\sz\linewidth]{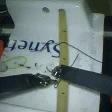}
	& \includegraphics[height=\sz\linewidth]{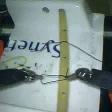}
	& \includegraphics[height=\sz\linewidth]{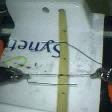}
	& \includegraphics[height=\sz\linewidth]{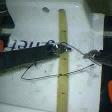} & &
	& \includegraphics[height=\sz\linewidth]{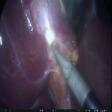}
	& \includegraphics[height=\sz\linewidth]{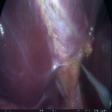}
	& \includegraphics[height=\sz\linewidth]{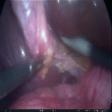}
	& \includegraphics[height=\sz\linewidth]{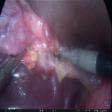} \\
	
	\includegraphics[height=\sz\linewidth]{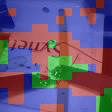}
	& \includegraphics[height=\sz\linewidth]{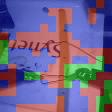}
	& \includegraphics[height=\sz\linewidth]{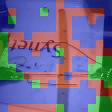}
	& \includegraphics[height=\sz\linewidth]{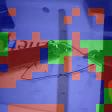} & &
	& \includegraphics[height=\sz\linewidth]{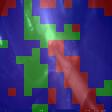}
	& \includegraphics[height=\sz\linewidth]{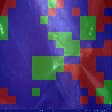}
	& \includegraphics[height=\sz\linewidth]{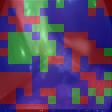}
	& \includegraphics[height=\sz\linewidth]{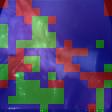} \\
    \end{tabular}
	\caption{Visualization of the assignment maps that display the correspondence between video parts and different semantic groups. }
	\label{fig_assignment}
\end{figure*}

%% file: tex_images/vis_htmpregu_gradcam.tex
\begin{figure*}[b!]
\centering
\newcommand{\sz}{0.2}
\renewcommand{\arraystretch}{0.9}
\begin{minipage}{.48\textwidth}
    \centering
	\begin{tabular}{cccc}
	
	\multicolumn{4}{l}{\scriptsize{Frames}} \\
	\includegraphics[height=\sz\linewidth]{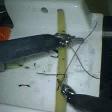} &
	\includegraphics[height=\sz\linewidth]{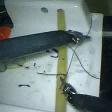} &
	\includegraphics[height=\sz\linewidth]{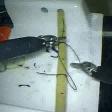} &
	\includegraphics[height=\sz\linewidth]{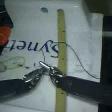} 
	\\
	
	\multicolumn{4}{l}{\scriptsize{Without SGM}} \\
	\includegraphics[height=\sz\linewidth]{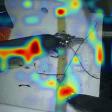} &
	\includegraphics[height=\sz\linewidth]{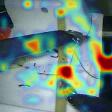} & 
	\includegraphics[height=\sz\linewidth]{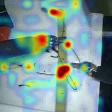} & 
	\includegraphics[height=\sz\linewidth]{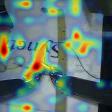} 
	\\
	
	\multicolumn{4}{l}{\scriptsize{With SGM}} \\
	\includegraphics[height=\sz\linewidth]{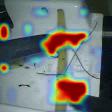} &
	\includegraphics[height=\sz\linewidth]{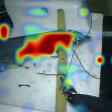} & 
	\includegraphics[height=\sz\linewidth]{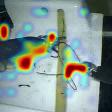} & 
	\includegraphics[height=\sz\linewidth]{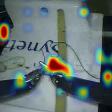} 
	\\
	
    \end{tabular}
	\caption{Visual explanations generated by Grad-CAM. SGM facilitates the concentration on the tools and discarding the unrelated background regions.}
	\label{fig:gradcam}
\end{minipage}
\hspace{0.5em}
\begin{minipage}{.48\textwidth}
	\centering
	\begin{tabular}{cccc}
	\multicolumn{4}{l}{\scriptsize{Without Position Supervision}} \\
	\includegraphics[height=\sz\linewidth]{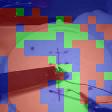} & 
	\includegraphics[height=\sz\linewidth]{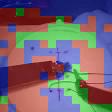} & 
	\includegraphics[height=\sz\linewidth]{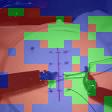} & 
	\includegraphics[height=\sz\linewidth]{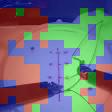} 
	\\
	
	\multicolumn{4}{l}{\scriptsize{Heatmaps for Position Supervision}} \\
	\includegraphics[height=\sz\linewidth]{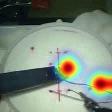} & 
    \includegraphics[height=\sz\linewidth]{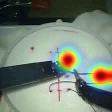} & 
	\includegraphics[height=\sz\linewidth]{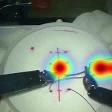} & 
	\includegraphics[height=\sz\linewidth]{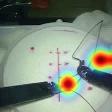} \\
	
	\multicolumn{4}{l}{\scriptsize{With Position Supervision}} \\
	\includegraphics[height=\sz\linewidth]{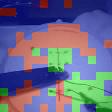} & 
	\includegraphics[height=\sz\linewidth]{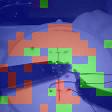} & 
	\includegraphics[height=\sz\linewidth]{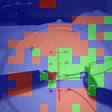} & 
	\includegraphics[height=\sz\linewidth]{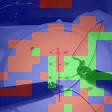} 
	\\
	
    \end{tabular}
	\caption{Improved semantic grouping results after using the tool position supervision. Green regions turn to focus on tool clips after using supervision.}
	\label{fig:heatmap_regu}
\end{minipage}
\end{figure*}

%% file: 05_discussion_conclusion.tex
\section{Conclusion}

In this paper, a novel framework called ViSA is proposed to predict the skill score from surgical videos by discovering and aggregating different semantic parts across spatiotemporal dimensions. 
The framework can achieve competitive performance on two datasets: JIGSAWS and HeiChole as well as support explicit supervision on the feature semantic grouping for performance improvement.


%% file: tex_tables/suptable_arch.tex
\begin{table*}[ht]
	\footnotesize
	\centering
	\renewcommand{\tabcolsep}{2pt}
	\renewcommand{\arraystretch}{1.2}

	\caption{\textbf{Network overview}. Details of major stages in ViSA. For the output sizes with multiple dimensions, the first dimension is the sequence length while the second dimension is the number of channels.}
	
    \begin{adjustbox}{width=\linewidth}
	\begin{tabular}{@{}l|l|l|l@{}}
		\toprule
		\textbf{Stage} & \textbf{Input (I)} \& \textbf{Output (O)} & \textbf{Implementation} & \textbf{Output Sizes} \\
		
		\midrule
		
        frames & N/A & N/A & $T\times3\times112\times112$ \\
        \hline
        
        \multirow{4}{*}{feature extracting} & 
        \multirow{2}{*}{I: frames} & R(2+1)D-18-Conv1 & $T\times64\times56\times56$ \\
        & & R(2+1)D-18-Conv2\_x & $T\times64\times56\times56$ \\
        & \multirow{2}{*}{O: feature maps} & R(2+1)D-18-Conv3\_x & $\frac{T}{2}\times128\times28\times28$ \\
        & & R(2+1)D-18-Conv4\_x & $\frac{T}{4}\times256\times14\times14$ \\
        \hline
        
        \multirow{2}{*}{semantic grouping} & I: feature maps & \multirow{2}{*}{-} & \multirow{2}{*}{$\frac{T}{4}\times256\times K\times1$} \\
        & O: group features & & \\
        \hline
        
        \multirow{2}{*}{group context modeling} & I: group features & Basic Residual $1\times1$ Conv Block & \multirow{2}{*}{$\frac{T}{4}\times256\times K$} \\
        & O: group contextual features & $K\times$ Bidirectional LSTMs (1 layer) & \\
        \hline
        
        \multirow{2}{*}{scene feature generating} & I: feature maps & \multirow{2}{*}{Average Pooling over Space} & \multirow{2}{*}{$\frac{T}{4}\times256\times 1\times1$} \\
        & O: scene features & & \\
        \hline
		
		\multirow{2}{*}{scene context modeling} & I: scene features & Basic Residual $1\times1$ Conv Block & \multirow{2}{*}{$\frac{T}{4}\times256\times 1$} \\
        & O: scene contextual features & $1\times$ Bidirectional LSTMs (1 layer) & \\
        \hline
        
        concatenating \& & I: group \& scene contextual & Concatenation & $\frac{T}{4}\times256\times(K+1)$ \\
        reshaping & features O: contextual features & Reshaping & $\frac{T}{4}\times256(K+1)$ \\
        \hline
        
        \multirow{2}{*}{temporal pooling} & I: contextual features & \multirow{2}{*}{Average Pooling over Time} & \multirow{2}{*}{$256(K+1)$} \\
        & O: global contextual features & & \\
        \hline
        
        \multirow{2}{*}{score regressing} & I: global contextual features & 1 \multirow{2}{*}{Fully Connected Layer} & \multirow{2}{*}{1} \\
        & O: skill score & & \\
        
		\bottomrule
	\end{tabular}
    \end{adjustbox}
	\label{tab:net}
\end{table*}

%% file: tex_tables/subtable_jigsaws_variance.tex
\begin{table*}[t]
\centering
\caption{Standard Deviations of the Spearman's Rank Correlation (Corr)  results on the JIGSAWS dataset.}
\label{tab:jigsaws_variance}
\begin{adjustbox}{width=\textwidth}
\setlength{\tabcolsep}{4pt}
\renewcommand{\arraystretch}{1.2}
\begin{tabular}{@{}ll|ccc|ccc|ccc|ccc@{}}
\toprule
\multirow{3}{*}{Method} & \multirow{3}{*}{Metric} & \multicolumn{12}{c}{Task \& Scheme} \\ 
\cmidrule{3-14} 
 &  & \multicolumn{3}{c}{KT} & \multicolumn{3}{c}{NP} & \multicolumn{3}{c}{SU} & \multicolumn{3}{c}{Across} 
\\ 
 \cmidrule(lr){3-5} \cmidrule(lr){6-8} \cmidrule(lr){9-11} \cmidrule(lr){12-14} 
 &  & LOSO & LOUO & 4-Fold & LOSO & LOUO & 4-Fold & LOSO & LOUO & 4-Fold & LOSO & LOUO & 4-Fold \\  
\midrule

\multirow{1}{*}{ViSA} 
 & Corr & 0.018 & 0.064 & 0.021 & 0.013 & 0.131 & 0.063 & 0.018 & 0.057 & 0.041 & 0.008 & 0.035 & 0.013 \\

\bottomrule
\end{tabular}
\end{adjustbox}
\end{table*}

%% file: tex_images/supvis_numpart.tex
\begin{figure}[t!]
    \centering
    \begin{subfigure}[b]{\textwidth}
        \hfill\includegraphics[width=\textwidth]{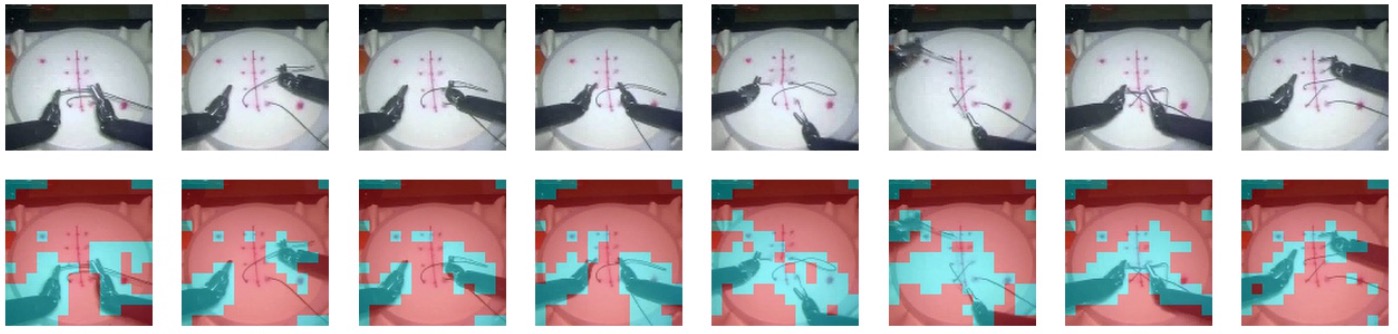}\put(-355,5){\rotatebox{90}{\scriptsize (a) K=2 \hspace{1em} Frames}}
     \end{subfigure}
     
     \begin{subfigure}[b]{\textwidth}
        \hfill\includegraphics[width=\textwidth,trim=0 0 0 160,clip]{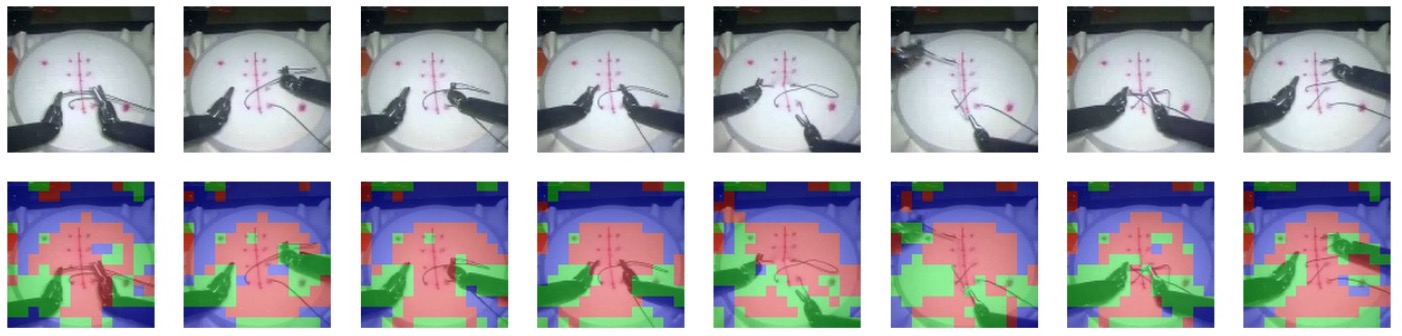}\put(-355,5){\rotatebox{90}{\scriptsize (b) K=3}}
     \end{subfigure}
     
     \begin{subfigure}[b]{\textwidth}
        \hfill\includegraphics[width=\textwidth,trim=0 0 0 160,clip]{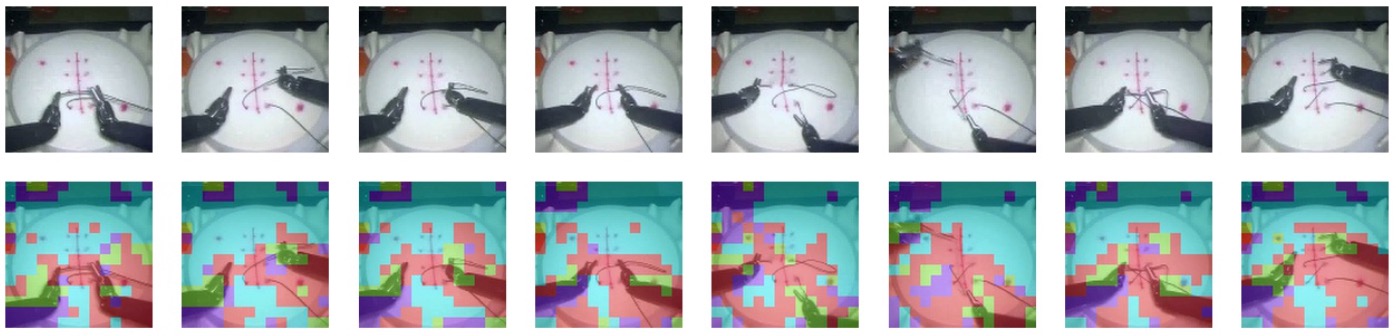}\put(-355,5){\rotatebox{90}{\scriptsize (c) K=4}}
     \end{subfigure}
     \caption{Visualization of the assignment maps generated by setting different numbers of semantic groups (K) and visualized with pseudo colors.}
     \label{fig:numparts}
\end{figure}

%% file: tex_images/supvis_r2d.tex
\begin{figure*}[t!]
	\centering
	\footnotesize
	\renewcommand{\tabcolsep}{1pt}
	\newcommand{\sz}{0.12}
	\renewcommand{\arraystretch}{1.2}
	\newcommand{\hlinevspace}{\hline\vspace{-7pt}}
	\begin{tabular}{cccccccc}
	
	\includegraphics[height=\sz\linewidth]{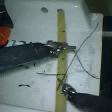} &
	\includegraphics[height=\sz\linewidth]{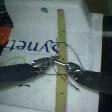} &
	\includegraphics[height=\sz\linewidth]{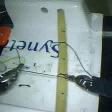} &
	\includegraphics[height=\sz\linewidth]{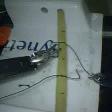} &
	\includegraphics[height=\sz\linewidth]{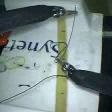} &
	\includegraphics[height=\sz\linewidth]{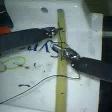} &
	\includegraphics[height=\sz\linewidth]{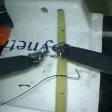} &
	\includegraphics[height=\sz\linewidth]{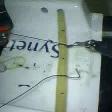} \\
	
	\includegraphics[height=\sz\linewidth]{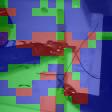} &
	\includegraphics[height=\sz\linewidth]{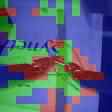} &
	\includegraphics[height=\sz\linewidth]{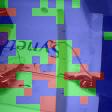} &
	\includegraphics[height=\sz\linewidth]{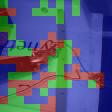} &
	\includegraphics[height=\sz\linewidth]{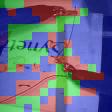} &
	\includegraphics[height=\sz\linewidth]{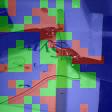} &
	\includegraphics[height=\sz\linewidth]{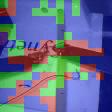} &
	\includegraphics[height=\sz\linewidth]{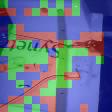} \\
    \end{tabular}
    \vspace{-1em}
	\caption{Visualization of the assignment maps generated by the framework extracting and aggregating 2D CNN features. The semantic grouping module can also work well with R2D and discover video parts with different semantics.}
	\label{fig:r2d}
	\vspace{-1em}
\end{figure*}

%% file: tex_tables/res_tasd_bsl.tex
\begin{table}[t]
\centering
\caption{Quantitative comparison results against baseline methods on TASD-2 for synchronised diving. We report and compare the Spearman's Correlations on the defined validation sets. \textbf{V}: video frames; \textbf{P}: body poses.}
\label{tab:tasd_baseline}
\begin{adjustbox}{width=0.45\linewidth}
\small
\begin{tabular}{@{}llccc@{}}
\toprule
\multirow{2}{*}{Input} & \multirow{2}{*}{Method} & \multicolumn{2}{c}{Task} \\ 
\cmidrule{3-5} 
 &  & 3m & 10m & Average \\ 
\midrule

\multirow{2}{*}{V}
 & C3D-LSTM~\cite{aqa_olympic} & -0.14 & 0.01 & -0.07 \\
 & I3D-SVR~\cite{aqa_asymmetric} & 0.84 & 0.83 & 0.83 \\
\midrule

\multirow{3}{*}{V+P}
 & JR-GCN~\cite{aqa_jointgraph} & 0.89 & 0.81 & 0.86 \\
 & AIM (Single-task)~\cite{aqa_asymmetric} & 0.89 & 0.85 & 0.87 \\
 & AIM (Multi-task)~\cite{aqa_asymmetric} & 0.92 & 0.85 & 0.89 \\
\midrule

\multirow{1}{*}{V}
 & ViSA & \textbf{0.97} & \textbf{0.95} & \textbf{0.96} \\

\bottomrule
\end{tabular}
\end{adjustbox}

    
        
        
        
        
        
\end{table}

%% file: tex_images/supvis_tasd_numpart.tex
\begin{figure*}[ht]
	\centering
	\footnotesize
	\renewcommand{\tabcolsep}{1pt}
	\renewcommand{\arraystretch}{1.2}
	\newcommand{\sz}{0.12}
	\newcommand{\hlinevspace}{\hline\vspace{-7pt}}
	\begin{tabular}{cccccccc}
	
	\includegraphics[height=\sz\linewidth]{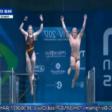} &
	\includegraphics[height=\sz\linewidth]{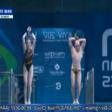} &
	\includegraphics[height=\sz\linewidth]{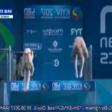} &
	\includegraphics[height=\sz\linewidth]{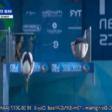} &
	\includegraphics[height=\sz\linewidth]{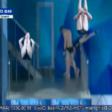} &
	\includegraphics[height=\sz\linewidth]{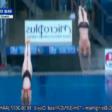} &
	\includegraphics[height=\sz\linewidth]{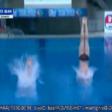} &
	\includegraphics[height=\sz\linewidth]{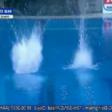} 
	\put(-355,5){\rotatebox{90}{\scriptsize Frames}} 
	\\
	
	\includegraphics[height=\sz\linewidth]{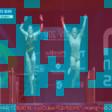} &
	\includegraphics[height=\sz\linewidth]{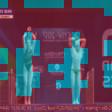} &
	\includegraphics[height=\sz\linewidth]{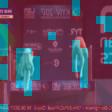} & 
	\includegraphics[height=\sz\linewidth]{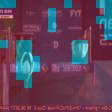} &
	\includegraphics[height=\sz\linewidth]{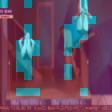} & 
	\includegraphics[height=\sz\linewidth]{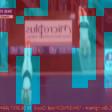} &
	\includegraphics[height=\sz\linewidth]{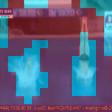} &
	\includegraphics[height=\sz\linewidth]{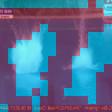} 
	\put(-355,10){\rotatebox{90}{\scriptsize K=2}}
	\\
	
	\includegraphics[height=\sz\linewidth]{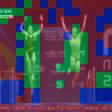} &
	\includegraphics[height=\sz\linewidth]{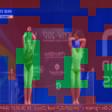} &
	\includegraphics[height=\sz\linewidth]{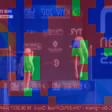} &
	\includegraphics[height=\sz\linewidth]{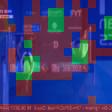} &
	\includegraphics[height=\sz\linewidth]{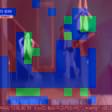} &
	\includegraphics[height=\sz\linewidth]{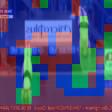} &
	\includegraphics[height=\sz\linewidth]{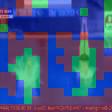} &
	\includegraphics[height=\sz\linewidth]{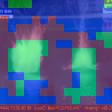} 
	\put(-355,10){\rotatebox{90}{\scriptsize K=3}} 
	
	
    \end{tabular}
	\caption{Visualization of the assignment maps generated by setting different numbers of semantic groups on the TASD-2 diving dataset.}
	\label{fig:tasd_numparts}
\end{figure*}